\documentclass[conference]{IEEEtran}
\IEEEoverridecommandlockouts
\usepackage{cite}
\usepackage{caption}
\usepackage{subcaption}
\usepackage{amsmath,amssymb,amsfonts}
\usepackage{algorithmic}
\usepackage[pdftex]{graphicx}
\usepackage{graphicx}
\usepackage{textcomp}
\usepackage{tabularx}
\usepackage{multirow}
\usepackage{multicol}
\usepackage{booktabs}
\usepackage{listings}
\usepackage{float}
\usepackage{mathtools}
\usepackage{xcolor}
\def\BibTeX{{\rm B\kern-.05em{\sc i\kern-.025em b}\kern-.08em
    T\kern-.1667em\lower.7ex\hbox{E}\kern-.125emX}}

\usepackage{siunitx}
\usepackage{caption}
\usepackage{subcaption}
\begin{document}
\setlength{\parskip}{0pt}
\title{Quantifying Demonstration Quality for Robot Learning and Generalization}

\author{Maram Sakr$^{*1,2}$, Zexi Jesse Li$^1$, H. F. Machiel Van der Loos$^1$,
 Dana Kuli{\'c}$^2$, and Elizabeth A. Croft$^2$

\thanks{$^{1}$Mechanical Engineering, University of British Columbia.}
\thanks{$^{2}$Faculty of Engineering, Monash University}
\thanks{$^{*}$Email: maram.sakr@ubc.ca}}

\maketitle
\thispagestyle{empty}
\pagestyle{empty}

\begin{abstract}
Learning from Demonstration (LfD) seeks to democratize robotics by enabling diverse end-users to teach robots to perform a task by providing demonstrations. However, most LfD techniques assume users provide optimal demonstrations. This is not always the case in real applications where users are likely to provide demonstrations of varying quality, that may change with expertise and other factors. Demonstration quality plays a crucial role in robot learning and generalization. Hence, it is important to quantify the quality of the provided demonstrations before using them for robot learning. In this paper, we propose quantifying the quality of the demonstrations based on how well they perform in the learned task. We hypothesize that task performance can give an indication of the generalization performance on similar tasks. The proposed approach is validated in a user study (N = 27). Users with different robotics expertise levels were recruited to teach a PR2 robot a generic task (pressing a button) under different task constraints. They taught the robot in two sessions on two different days to capture their teaching behaviour across sessions. The task performance was utilized to classify the provided demonstrations into high-quality and low-quality sets. The results show a significant Pearson correlation coefficient $(\rho = 0.85, p < 0.0001)$ between the task performance and generalization performance across all participants. We also found that users clustered into two groups: Users who provided high-quality demonstrations from the first session, assigned to the \textit{fast-adapters} group, and users who provided low-quality demonstrations in the first session and then improved with practice, assigned to the \textit{slow-adapters} group. These results highlight the importance of quantifying demonstration quality, which can be indicative of the adaptation level of the user to the task.

\end{abstract}


\section{INTRODUCTION}
        \label{sec:introduction}
        As robots enter the human environment to assist people in their daily lives, the ability of everyday users, who do not have any robotics or programming background, to work with these robots will soon become essential. A common approach to allow ordinary people to program a robot is Learning from Demonstration (LfD). LfD is a paradigm that allows robots to perform tasks after observing a teacher performing them without explicit programming \cite{argall2009survey}.

One of the key elements for deploying LfD in the real world is ensuring fast learning and generalization. This means that the robot has the ability to learn the taught task quickly and apply the learned knowledge to new circumstances. It is not practical to reprogram the robot for every small change in the task and/or environment. To design a generalizable learning system, we need to consider both the human teacher and the robot learner as active stakeholders who have a role in improving efficiency \cite{calinon2007teacher}. Most of the literature to date has focused on the robot side, that is, on advancing techniques to boost robot learning efficiency \cite{osa2018algorithmic}. Much less attention has been paid to the human teacher's role in this process.

As noted in \cite{argall2009survey}, a robot’s performance depends heavily on the quality of demonstrations provided. Hence, it is crucial to define and measure demonstrations quality when teaching a robot. There are different factors that affect demonstrations quality, here we focus on the provider of the demonstrations. The human teacher's adaptation to the task and the demonstration interface may be varying, and is related to their previous expertise. Kobak and Mehring \cite{kobak2012adaptation} highlighted that after practice with motor tasks sharing structural similarities, new tasks of the same type can be learned faster. Thus, users' stated robotics expertise may not be sufficient to judge their expertise level and their provided demonstrations quality~\cite{everson1997metacognitive}. This necessitates the need for a quantitative approach for quantifying demonstration quality rather than simply relying on the stated expertise level of the users. 


\begin{figure}[t]
  \begin{subfigure}[b]{0.49\columnwidth}
    \includegraphics[width=\linewidth]{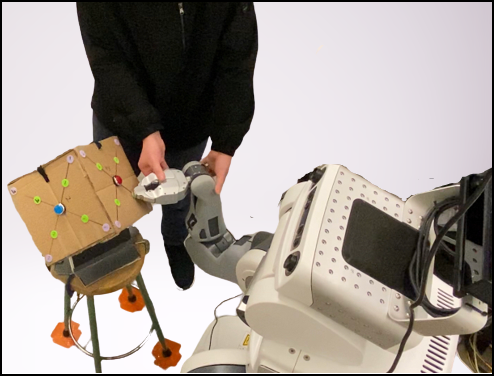}
    \caption{(a) Low-constraint task}
  \end{subfigure}
  \hfill 
  \begin{subfigure}[b]{0.49\columnwidth}
    \includegraphics[width=\linewidth]{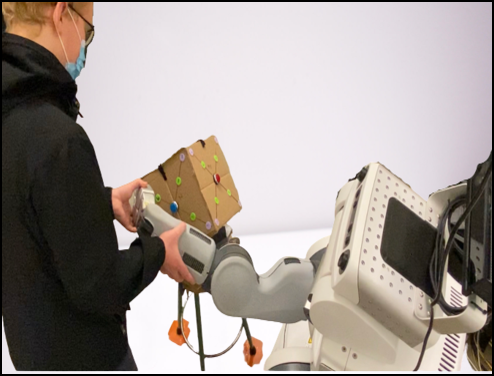}
    \caption{(b) High-constraint task}
  \end{subfigure}
  \caption{Human demonstrators kinesthetically teach a robot to press a button on a box on (a) a low-constraint face close to the robot and (b) a high-constraint face such that the robot must be maneuvered around the box in a tight space.}
  \label{fig:pr2}
\end{figure}
The term "poor quality demonstrations" has been used loosely in the literature to refer to different issues in the data itself. These issues include undesired motions~\cite{osa2018algorithmic}, failed demonstrations~\cite{grollman2011donut}, and ambiguous demonstrations~\cite{sena2020quantifying}. However, the explicit definition and measurements for the \textit{demonstration quality} is still an open question~\cite{osa2018algorithmic}.

\begin{figure*}[t]
\centering
\includegraphics[width=0.85\textwidth]{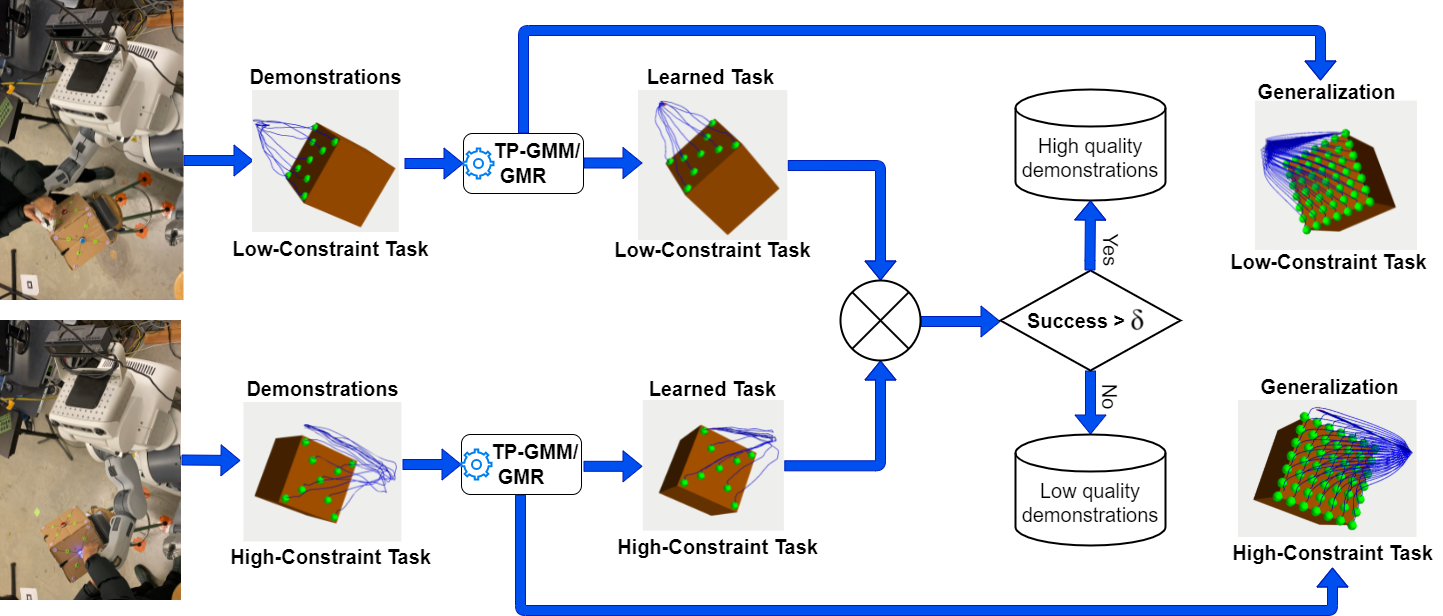}
\caption{The proposed approach for assessing demonstrations using learned task performance in a task with two levels of constraints.}
\label{fig:approach}
\end{figure*}
In this paper, we define the quality of the provided demonstrations based on how well the learning algorithm reproduces the task, regardless of the demonstrator's stated expertise level. This will be done by assessing the learning algorithm on the same task set as demonstrations, then on a generalized version of the demonstrated task.

\section{BACKGROUND AND RELATED WORK}
        \label{sec:related_works}
        
A number of researchers have attempted to assess the quality of the provided demonstrations. Ureche and Billard~\cite{pais2015metrics} proposed three metrics for assessing human demonstrations of bi-manual tasks. These metrics are the ability to maneuver the tool, the consistency in teaching and the degree of coordination between the two arms. The main limitation of these metrics is that the consistency in teaching does not directly reflect the quality. The users may provide consistent demonstrations with the same errors due to their lack of understanding of \textit{what} and \textit{how} the robot learns during task demonstration~\cite{cakmak2014eliciting, hellstrom2018understandable}.

Kaiser et al. ~\cite{kaiser1995obtaining} listed different sources of sub-optimality in the human demonstrations including: i) unnecessary actions that do not contribute to the final goal; ii) incorrect actions that negatively affect the usefulness of the demonstrations; and, iii) unmotivated actions by the human teacher that are measured by sensors that are not available to the robot. To avoid unmotivated actions, teaching methods that record the demonstration data from robot's body can be used, such as teleoperation and kinesthetic teaching~\cite{billard2008survey}. To avoid unnecessary actions, a metric was proposed to detect such actions leaving it is up to the experimenter to include demonstrations with unnecessary actions or ignore them. In ~\cite{kaiser1995obtaining}, the authors noted that it was difficult to distinguish incorrect actions from correct ones.

Fischer et al. ~\cite{fischer2016comparison} compared between different teaching interfaces in robot learning from demonstration. They found three main errors commonly committed by the users: i) applying to much pressure to the gripper's fingers, ii) moving into a singularity, and iii) moving into a self-collision. However, they did not study the effect of these errors on robot learning and generalization. Jaquier et al. ~\cite{Jaquier2020geometry} proposed a framework that allows robots to learn and reproduce the joint space trajectories with their particular manipulability indices. 

Recently, Sena and Howard~\cite{sena2020quantifying} identified three main issues with poor quality demonstrations. These issues are undemonstrated states, ambiguous demonstrations and unsuccessful demonstrations. Undemonstrated states refer to the states in which the robot can not perform the task because they have insufficient demonstrations to permit generalization to these states. Ambiguous demonstrations refer to the demonstrations that offer little or no new information to the learning model; this can happen if the demonstrations are very similar. Unsuccessful demonstrations refer to the demonstrations that do not achieve the task goal~\cite{grollman2011donut}. Sena and Howard propose two metrics for measuring demonstration sufficiency: i) teaching efficacy, that is, how much the robot can generalize over the entire task space; and ii) teaching efficiency, that is, teaching efficacy normalized by the number of provided demonstrations. Their ultimate goal is to achieve the highest efficacy given the fewest possible demonstrations. These two metrics are used to solve one of the main issues when collecting demonstrations from human teachers, namely data sparsity~\cite{argall2009survey}. However, these metrics do not address the \textit{quality} of the provided demonstrations that affect robot learning and generalization~\cite{Jaquier2020geometry}.

In this paper, we propose relating the quality of the provided demonstrations to the task learning and generalization they achieve. Learning is defined with respect to the task space covered by the demonstrations, and generalization is defined with respect to the task space adjacent to the space spanned by the demonstrations. We hypothesize that if demonstrations lead to high performance in task learning, they will also achieve high generalization performance. This is valid if the learning algorithm is compatible with the data type and size, and the test conditions in the generalization are from the same distribution as the conditions for the demonstrations ~\cite{sugiyama2015introduction}.
        

\section{PROPOSED APPROACH}
        \label{sec:approach}

Since task performance and generalization are the most important goals in any LfD system~\cite{osa2018algorithmic}, we propose using the task performance and generalization performance to judge the quality of the input demonstrations. 
The provided demonstrations are fitted into a learning model. This model will be evaluated on the same task set covered by the demonstrations and on adjacent task set to evaluate its generalizability. If the learning model performs the task on the same demonstrations' conditions with a success rate higher than a pre-defined threshold $(\delta)$, the provided demonstrations will be considered as high-quality ones. Otherwise, the provided demonstrations will be considered low-quality ones, as shown in Figure~\ref{fig:approach}. After classifying the provided demonstrations, the learning model will be tested on a new task set to evaluate its generalizability. This will be done to evaluate our hypothesis that the demonstrations that have high performance on the demonstrated task set will also have high generalization performance and vice versa. 
 
Several factors contribute to the generalization ability of any LfD system. For instance, the number of the provided demonstrations, their distribution over the task space, the demonstrator's expertise level, and the complexity of the task, among others. In this paper, we focus on the quality of the provided demonstrations over two levels of task complexity while fixing all other factors (i.e., the number of provided demonstrations and the task space). 
\vspace{-7pt}
\subsection{Robot Task Learning}
\label{sec:learning}
The learning process starts with the user providing a demonstration set consisting of \(M\) demonstrations. Each demonstration contains $T_m$ state data points. $T_m$ is assumed to be equal for each demonstration in the data set. If the demonstrations are not equal in size, they can be aligned using any alignment algorithm (e.g., dynamic time warping). Here we used the state measurements \(\xi_n = \left(t_n, \mathbf{x}_n^T, \boldsymbol{\epsilon}_n^T \right) \in \mathbb{R}^8 \), which involves the time \(t\), the end effector position \(\mathbf{x}_n^T\), and the end effector orientation represented by quaternions \(\boldsymbol{\epsilon}_n^T\).

A task parameterized Gaussian Mixture Model (TP-GMM) combined with Gaussian Mixture Regression (GMR) is used for task learning. TP-GMMs have been extensively used in the LfD literature~\cite{pervez2018learning, osa2018algorithmic}, and provide good generalization using a limited set of demonstrations. TP-GMM models a task using task parameters defined by a sequence of coordinate frames. In a \(D\)-dimensional space, each task parameter/coordinate frame is given by an $A \in \mathbb{R}^{D \times D}$ matrix indicating its orientation and a $\boldsymbol{b} \in \mathbb{R}^D$ vector indicating its origin, relative to the global frame. A \(K\)-component mixture model is fitted to the data in each local frame of reference. Each GMM is described by $\left(\pi_k, \mathbf{\mu}^{(j)}_k, \mathbf{\Sigma}^{(j)}_k\right)$, referring to the prior probabilities, means, and covariance matrices for each component \(k\) in frame \(j\), respectively. An expectation-maximization (EM) algorithm is used to estimate these parameters, with the constraint that data points from different frames of reference must arise from the same source. To use the local models for trajectory generation, they must be projected back into the global frame of reference and then combined into one global model. This is achieved through a linear transformation of the local models with their respective task parameters, followed by a product of Gaussians. A new GMM with components $\left\{\pi_k, \mathbf{\mu}_{k,t}, \mathbf{\Sigma}_{k,t} \right\}_{k=1}^K$ at time $t$ in the global frame $\left\{O \right\}$ can be computed as:
\begin{equation}
    \mathcal{N}\left(\boldsymbol{\mu}_{k,t}, \boldsymbol{\Sigma}_{k,t} \right) \propto \prod^J_{j=1} \mathcal{N}\left(\mathbf{A}_t^{(j)} \boldsymbol{\mu}_k^{(j)} + \mathbf{b}_t^{(j)}, \mathbf{A}_t^{(j)} \boldsymbol{\Sigma}_k^{(j)} (\mathbf{A}_t^{(j)})^T \right)
\end{equation}

Then, GMR can be used to obtain the next trajectory point. This procedure is repeated for each time step in the trajectory. Calinon \cite{calinon2016tutorial} provides more in-depth detail for this approach.


\subsection{Joint Space Learning}
\label{sec:IK}
Given the task space trajectory from TP-GMM represented as task position $x(t)$ and task velocity $\dot{x}(t)$, the goal is to find a feasible joint space trajectory as joint position $q(t)$ and joint velocity $\dot{q}(t)$ that reproduce the given trajectory. The differential kinematics equation establishes a linear mapping between joint space velocities and task space velocities, and it can then be utilized to solve for joint velocities. 

To avoid kinematic singularities, we used a singularity-robust (SR) inverse \cite{nakamura1986inverse}, also known as a damped pseudoinverse \cite{maciejewski1990dealing}. For a 7-DoF redundant manipulator, a nonempty null space exists due to the excess of input space relative to the manipulable space. We used a common method of including the null space in a solution with the formulation in \cite{liegeois1977automatic} as follows:

\begin{equation}
\label{nullspace}
\dot{q} = J^*(q)\dot{x} + (I - J^\dagger(q)J(q))\pi(q)
\end{equation}

\noindent $J^*$ is the SR inverse Jacobian, the pseudoinverse $J^\dagger$ can be computed as $J^T (JJ^T)^{-1}$, the matrix $(I - J^\dagger(q)J(q))$ is the null space projection operator, and $\pi(q)$ is the null space policy. $\pi(q)$ can be used to control motion in  joint space without affecting the task-space motion. Since our goal is to mimic human demonstrations in both Cartesian and joint spaces, we used the closest demonstration as the null space policy $\pi(q)$ for the generated trajectory. 

The open-loop solutions of joint variables through numerical integration unavoidably leads to drift and accumulated task space errors. To overcome these drawbacks, the closed-loop inverse kinematics (CLIK) algorithm with error feedback is utilized. The CLIK algorithm adds a proportional feedback loop that adds task space error to the desired task space velocity. The CLIK algorithm can be expressed by the following equation:
\begin{equation}
\label{nullspace}
\dot{q} = J^*(q)(\dot{x}_d + k_p(x_d - x)) + (I - J^\dagger(q)J(q))\pi(q)
\end{equation}

\noindent where $k_p$ is a symmetric positive definite matrix, and the choice of $k_p$ guarantees that the error uniformly converges to zero. $\dot{x}_d$ is the desired velocity in task space, and $(x_d - x)$ is the position error in task space.

Finally, the joint values output from the CLIK algorithm are checked to see if any joint is close to its limits. If so, the CLIK calculation will be redone. This procedure is repeated until valid joint values are achieved.

\section{EXPERIMENTAL DESIGN}
        \label{sec:experiment}
        \subsection{Robot Platform}
The robot platform used in this work is the PR2 (Willow Garage, Personal Robot 2), a mobile manipulator with two 7-DoF arms and an omnidirectional base. The passive spring counterbalance system in PR2’s arms provides gravity-compensation, giving users the ability to kinesthetically move the robot's arms within their kinematic range. Each arm has a 1-DoF under-actuated gripper. Its wrists have two axes of freedom. In this experiment, we only used the right arm in gravity compensation mode with the gripper closed. 

\subsection{Task Definition}
The exemplar task chosen is pressing a button; this is a general task for pressing a doorbell, elevator call button, pedestrian crossing button, etc. The task was chosen to be sufficiently generic that it does not require domain expertise, but does require practice with the robot to provide high-quality demonstrations. This task comprises both a constrained reaching task as well as fine control motion for pressing the button. Furthermore, the task was subject to two different levels of configuration and task-space constraints: low-constraint and high-constraint. This line of investigation is motivated by Fitt's law~\cite{fitts1954information} where the logarithm of the ratio of the target distance and the target width are used to represent the difficulty level. Here, we used target distance and constraint level as the difficulty aspects.

Figure~\ref{fig:pr2} shows the experimental setup used for data collection. As shown in the figure, a cardboard box was fixed on one of its vertices such that all buttons are reachable by the robot gripper. Only two faces of the box were used in the data collection. On each face, buttons were placed in the centre (large green button), corners (purple foam markers), and at locations midway between the corners and the centre of each face (green foam markers). A total of nine goal positions for each face were used. Face-1 represents a low-constraint task as the robot can easily reach all the target points. Face-2 represents a high-constraint task as the participant needs to maneuver the robot arm around the box in a constrained space to reach the goal positions while avoiding self collisions and collisions with the box as shown in Figure~\ref{fig:pr2}-b.


\subsection{User Study}
We recruited participants for our user study through word of mouth, advertisements posted on university campus, and social media. A total of 27 participants (21 male, 6 female) with $\mu_{age} = 22$ with different robotics expertise levels ranging from no robotics experience to 6 years or more of experience were recruited. Prior to conducting the study, we obtained research ethics approval from the university's Behavioural Research Ethics Board (application ID H20-03740). We obtained informed consent from each participant before commencing the experiment.

The experiment was conducted in a two-session regime on two different days. This two-session format is motivated by the fact that practice is essential for motor skill learning~\cite{schmidt2008motor}. We were particularly interested to explore which aspects improved from the first session to the second and how this will be reflected in robot learning and generalization. Participants were scheduled ahead of time for one hour time slots. Upon arrival, they were brought to the experiment area and given an informed consent form to sign. The experimenter briefly stated the long-term goal of the research, and told the participants that their task will be to program new skills on the PR2. The experimenter did not give any instructions on how to program the PR2, rather participants were asked to learn by doing, in order to record their first-time (novice) interaction with the robot. The robot was set in gravity compensation mode. Participants were asked to hold the robot's right arm and physically guide it to press a target button (kinesthetic teaching). The right arm always started at the same position with elbow at 90° and the gripper pointed up (untucked position). The experimenter teleoperated the right arm via joystick to bring the arm back to the initial position after each demonstration. The participants started with guiding the robot to press the nine Face-1 buttons (low-constraint task) and then progressed to the nine Face-2 buttons (high-constraint task) with three trials for each face. The robot joint angles were recorded during each demonstration and saved as ROSbag files for offline analysis. A total of 54 demonstrations were collected in a session for each participant. At the end of the first session, the experimenter asked the participants to reflect on what they had learned and practiced in the first session on how to teach the robot and to schedule a second session on a different day. This procedure was motivated by the findings of Walker et al. ~\cite{walker2003sleep} that a night of sleep after training on motor skill significantly improves the skill level in a later retesting. In the second session, each participant repeated the same procedure as the first session. At the end of the second session, each participant answered a questionnaire\footnote{\url{https://forms.gle/TJ6ULsWVkV2dwUpY6}} about their previous experience in robotics, if any. In addition, some open questions about the aspects of their performance that they felt improved between the two sessions. 

\subsection{Performance Metrics}

To evaluate the collected demonstrations, a TP-GMM model was created using each trial's demonstrations as described in Section~\ref{sec:learning}. Then, the model was tested on the same demonstrated task set as well as 49 new target positions (generalization as shown in Figure~\ref{fig:approach}). Each face of the box in Figure ~\ref{fig:pr2} was discretized into a grid of 7x7 that gave a total of 49 new targets. The grid size was chosen based on the dimensions of the PR2 arm's tip and the box dimensions to avoid any overlap between targets. The arm tip's dimensions are (W = 2.1 cm, L = 2.2 cm, H = 3.5 cm), and the box is a cube of equally sized edges of 26 cm. We specified the target point as a sphere of 3 cm in diameter based on the centre button. The inverse kinematics (IK) procedure in Section~\ref{sec:IK} was used to compute joint-space trajectory for the learned Cartesian trajectory from TP-GMM. From the joint-space trajectory we can check whether the learned trajectory is a feasible and compute the collision rate of the robot with the experimental setup. 

The learned trajectory is considered successful if it reached the goal position (within the goal sphere) while avoiding self collisions and collisions with the box. To account for the non-zero size of the end effector tip, we consider the robot to have reached the goal if any point of the tip touches the goal sphere. To calculate the success rate for each trial, we divide the number of successful trajectories over the total number of the tested points (nine in the task performance test and 49 in the generalization test).

\subsection{Fast-adapter and Slow-adapter Users Definition}
The proposed approach in Figure~\ref{fig:approach} was utilized to classify the provided demonstrations in the two sessions into low-quality and high-quality ones. We found a group of users who provided high-quality demonstrations in both sessions, while others provide low-quality demonstrations in the first session and then after practice they provide high-quality ones in the second session. We call the first group of users \textit{fast-adapters} as they quickly adapt to the task and provide high-quality demonstrations consistently. The other group was called \textit{slow-adapters} as they need some time and practice to adapt to the task and provide high-quality demonstrations. 

\section{HYPOTHESES}
        \label{sec:hypotheses}
        We expect that task performance and generalization performance will be highly correlated. In addition, we expect that the user's performance will be improved with practice and that this improvement will be reflected in robot learning and generalization performance. With practice, we expect slow-adapters will converge to a strategy for teaching the robot how to do a task. This will be reflected in the improvement in the success rate from the first session to the second one. For fast-adapters, they already demonstrate good performance in the first session, so we expect that fast-adapters will not have a significant difference in their performance in the two sessions. Based on these expectations, we formulate the following hypotheses:
\begin{itemize}
    \item \textbf{H1}: There is a significant correlation between task performance and generalization performance. 
    \item \textbf{H2}: The task performance of fast-adapters' demonstrations is significantly higher than slow-adapters' demonstrations.
    \item \textbf{H3}: The task performance using second session demonstrations is significantly higher than using the first session ones for slow-adapters. 
    \item \textbf{H4}: The generalization performance of fast-adapters' demonstrations is significantly higher than slow-adapters' demonstrations.
    \item \textbf{H5}: The generalization performance using second session demonstrations is significantly higher than using the first session ones for slow-adapters. 
\end{itemize}

\section{RESULTS}
        \label{sec:results}

\begin{figure}[t]
\centering
\includegraphics[width=0.45\textwidth]{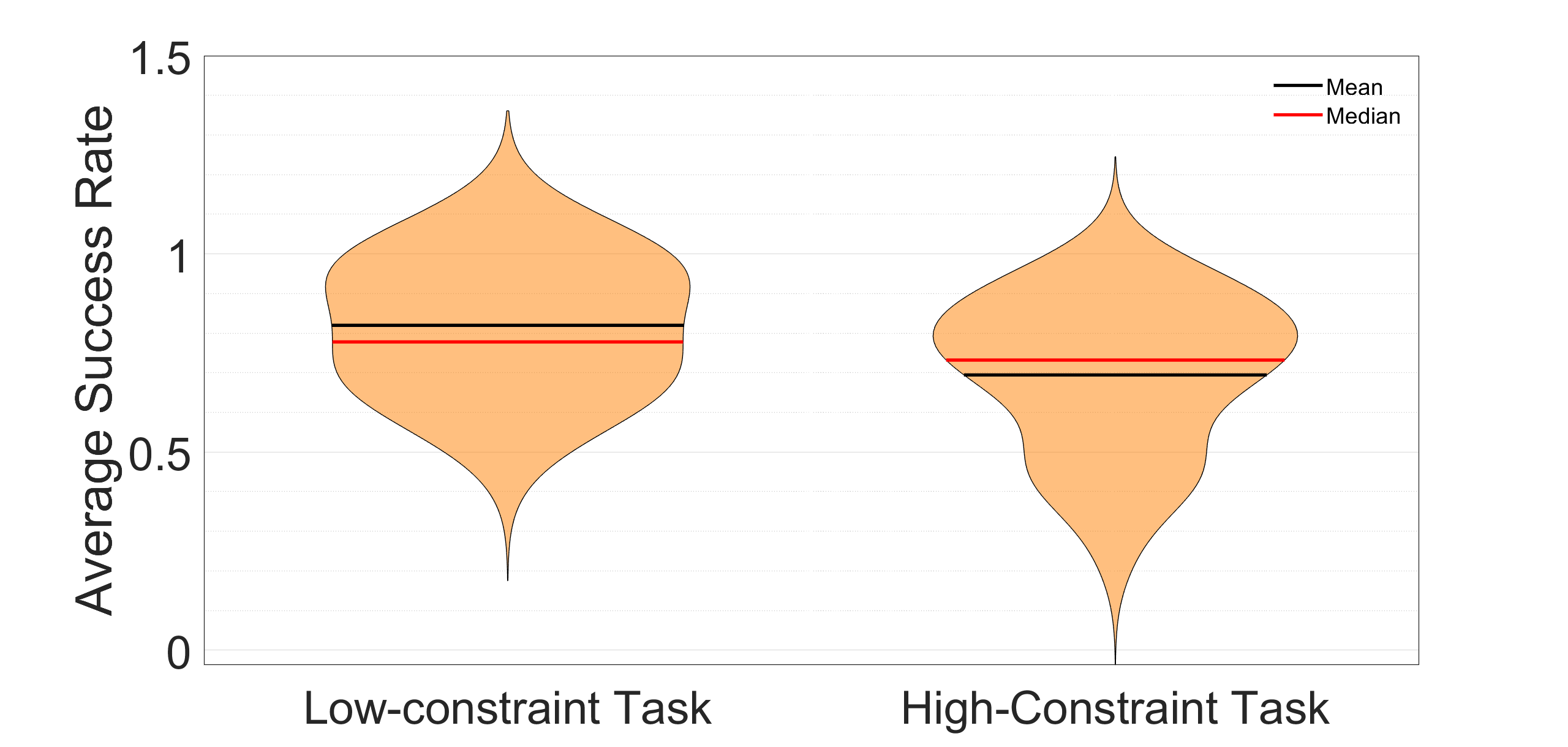}
\caption{Average success rate distribution of task performance of Session-1 on both task constraints.}
\label{fig:violin}
\end{figure}

In order to evaluate all hypotheses, a TP-GMM model was created using each trial's demonstrations and tested on the same demonstrated task set (task performance) as well as 49 new target positions (generalization performance). The Pearson correlation coefficient between task performance and generalization performance was calculated. We found a statistically significant positive correlation between task and generalization performance$(\rho = 0.85, p < 0.0001)$. Following the proposed approach in Figure~\ref{fig:approach}, we used task performance success rate to cluster the users into: fast-learners and slow-learners. To define the success rate threshold to categorize the users, we looked into the distribution of the average success rate of Session-1 trials for all participants as shown in Figure~\ref{fig:violin}. The violin figure shows a clear bi-modal distribution especially with high-constraint tasks. Since the mean and median of both task constraints are different, the threshold is defined as a value between them. A threshold of 80\% was chosen to categorize the users into: fast-learners and slow-learners. We found that 12 participants out of the 27 participants are fast-adapters with a success rate $>80\%$ for both low- and high- constraint tasks using the first session's demonstrations. Only six participants out of these 12 fast-adapters claimed previous robotics experience in the questionnaire, while four participants out of the 15 slow-adapters claimed previous experience in robotics. 
\subsection{Task Performance Analysis}
\begin{figure}[t]

\centering
\includegraphics[width=0.5\textwidth]{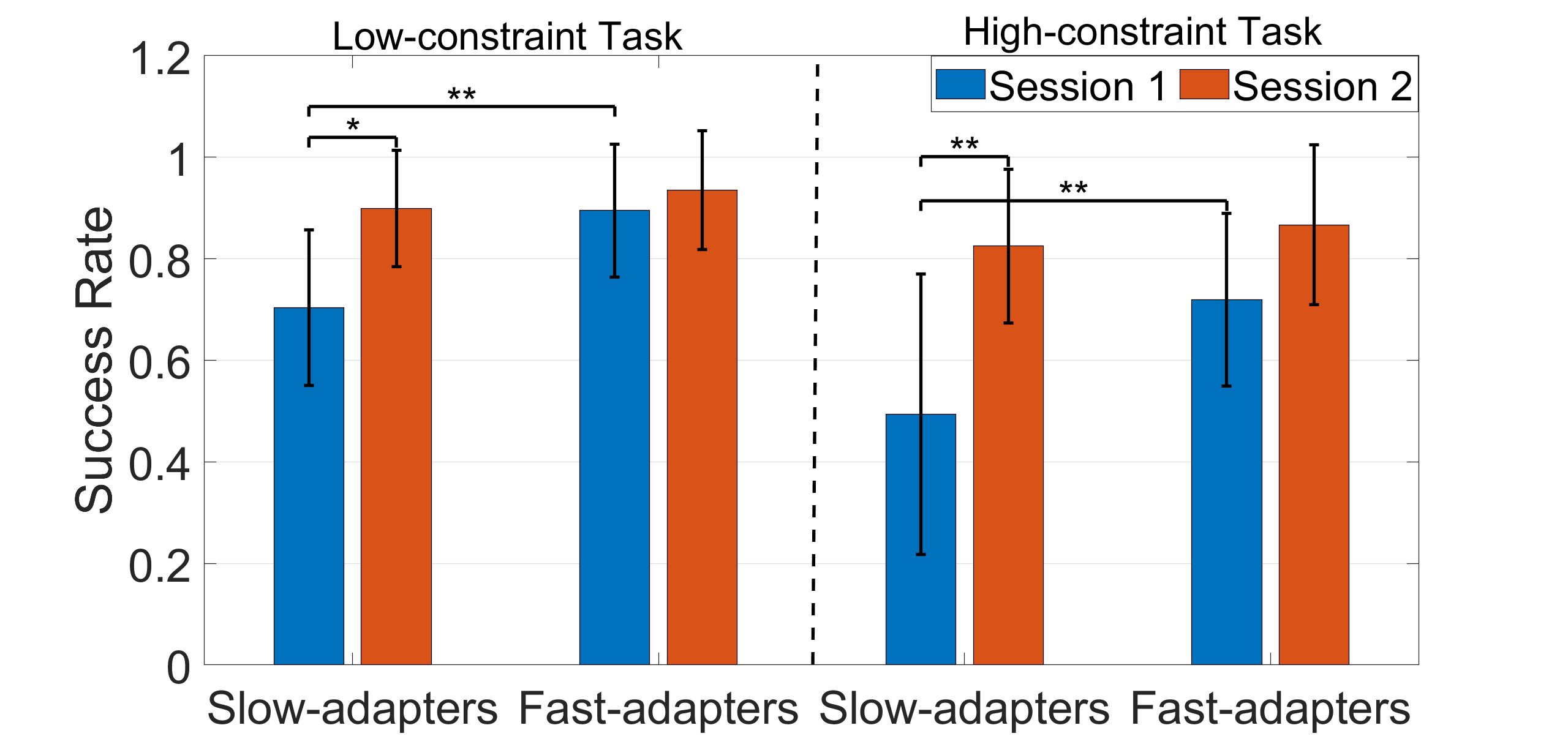}
\caption{Learned task success rate for both slow- and fast- adapters in the two sessions of two levels of constraints task.}
\vspace*{-3mm}
\label{fig:interpolation_means}
\end{figure}

Figure~\ref{fig:interpolation_means} shows the mean and standard deviation for success rate in task performance for both fast-adapters and slow-adapters in two sessions and two task constraints. Overall, Session-2 has a higher success rate with less variance than Session-1 for both fast- and slow-adapters in both task constraints. Furthermore, the improvement between the two sessions in the high-constraint tasks is higher than the improvement in the low-constraint tasks. Finally, there is a large difference between fast-adapters and slow-adapters in Session-1 while in Session-2 this difference decreased. 

A 2x2x2 mixed model ANOVA was conducted to investigate the impact of a) Adaptation level, b) Sessions and c) Task constraints on the learned task success rate. This rate was calculated by testing the learning model for each participant on the same nine demonstrated target points. Hence, the success score is the number of points at which the robot generates a successful trajectory without collisions over the total number of target points. The data were checked to be compatible with the relevant statistical assumptions. We found that six sets of data out of the eight we have do not follow a normal distribution using the Shapiro-Wilk test. ANOVA tests are noted as being robust to violations of normality~\cite{schmider2010really}, and considering that the excess kurtosis for the six groups is -0.525, 0.854, 7.561, 0.897, 0.538, and 0.522 (where a normal distribution would have an excess kurtosis value of zero), the violation is considered minor except for one case and the data are assumed to follow a normal distribution. The assumption of sphericity was not applicable as there were only two levels of all factors. Levene's test showed that all eight data sets do not violate the homogeneity of variances assumption. 

A significant main effect for sessions was obtained, $F(1,25) = 32.94, p<0.001, \eta_p^2 =0.568$ with confidence levels the success rate increased from $0.70 \pm 0.027$ in Session-1 to $0.88 \pm 0.018$ in Session-2. A significant main effect for adaptation level was obtained $F(1,25) = 2170.77, p<0.001, \eta_p^2 = 0.989$ with confidence levels the success rate increased from $0.73 \pm 0.023$ for slow-adapters to $0.85 \pm 0.025$ for fast-adapters. A significant main effect for task constraints was obtained, $F(1,25) = 15.26, p<0.001, \eta_p^2 =0.379$ with confidence levels the success rate decreased from $0.86 \pm 0.019$ in low-constraint tasks to $0.73 \pm 0.028$ in high-constraint tasks. In addition, the interaction between adaptation level and sessions showed a significant effect, $F(1,25) = 7.36, p = 0.012, \eta_p^2 = 0.228$ 
Examination of the marginal means indicated that slow-adapters have a notable improvement in their success rate from $0.60 \pm 0.037$ in Session-1 to $0.86 \pm 0.023$ in Session-2, while fast-adapters have a lower improvement from $0.81 \pm 0.041$ in Session-1 to $0.90 \pm 0.026$ in Session-2. 

The pairwise comparisons with Bonferroni correction showed that slow-adapters have significant improvement in their success rate from Session-1 to Session-2 $p<0.001$ while fast-adapters have marginal improvement from Session-1 to Session-2 $p=0.053$. Another interesting finding is that there is a significant difference between fast- and slow-adapters in Session-1 $p<0.001$ while there is no significant difference between them in Session-2. 

\begin{figure}[t]

\centering
\includegraphics[width=0.5\textwidth]{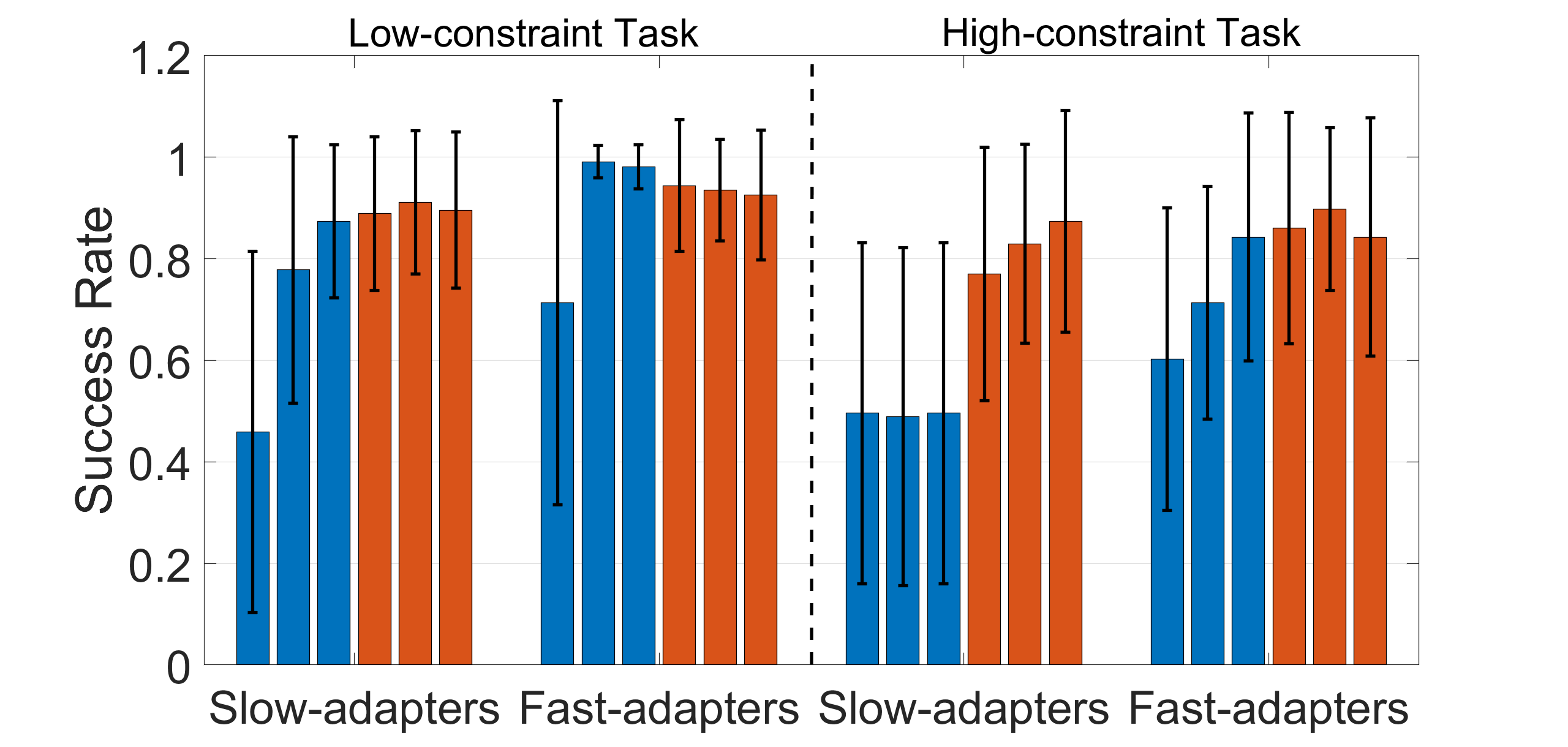}

\caption{Learned task success rate for both slow- and fast- adapters in all trials. Blue bars represents Session-1 trials and red bars represent Session-2 trials.}
\vspace*{-3mm}
\label{fig:interpolation_trials}
\end{figure}

Examining the individual trials of both sessions, we found that slow-adapters tend to take a longer time to reach high performance than the fast-adapters, as shown in Figure~\ref{fig:interpolation_trials}. Particularly in the high-constraint tasks, slow-adapters have a steady low performance in the first session's trials with $0.49 \pm 0.34$, then they gradually improve in the second session's trials till they reach $0.87 \pm 0.22$. On the other hand, fast-adapters gradually improve in Session-1 and almost converge to high performance at the end of their first session with $0.90 \pm 0.16$. For the low-constraint tasks, slow-adapters converge to their highest performance at the end of Session-1 with $0.87 \pm 0.15$ while fast-adapters converge to a higher performance of $0.99 \pm 0.032$ than slow-adapters right in the second trial of Session-1.

\subsection{Task Generalization Performance Analysis}
\begin{figure}[t]

\centering
\includegraphics[width=0.5\textwidth]{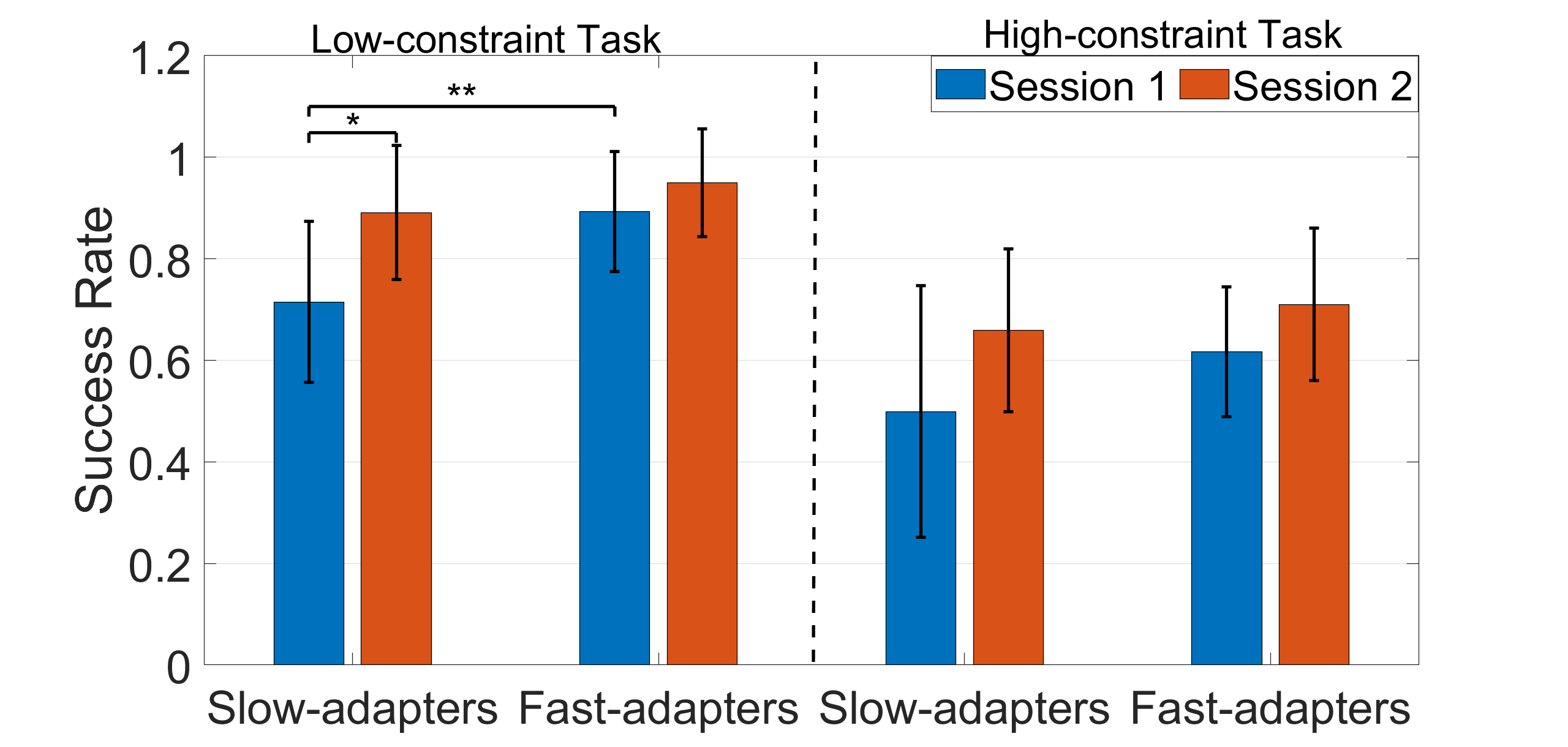}

\caption{Generalization success rate for both slow- and fast-adapters in the two sessions of two levels of constraints task.}
\vspace*{-3mm}
\label{fig:extrapolation_means}
\end{figure}

Figure~\ref{fig:extrapolation_means} shows the mean and standard deviation for success rate in task generalization for both fast- and slow-adapters for two sessions and two task constraints. Overall, it shows a similar trend as the task performance figure with Session-2 having a higher success rate with less variance than Session-1 for both fast- and slow-adapters in both task constraints. Also, there is a large difference between slow- and fast-adapters in Session-1 while in Session-2 this difference decreased. Finally, the low-constraint tasks shows a consistent success rate in both task learning and generalization performance while the high-constraint tasks has a degradation in the success rate in generalization performance compared to task learning.

A 2x2x2 mixed model ANOVA was conducted to investigate the impact of a) Adaptation level, b) Sessions and c) Task constraints on generalization success score. This score was calculated by testing the learning model for each participant on 49 new target points. Hence, the success score is the number of points at which the robot generates a successful trajectory without collisions over the total number of test target points. The data were checked to be compatible with the relevant statistical assumptions. We found that only three sets of data out of the eight did not follow a normal distribution using the Shapiro-Wilk test. The excess kurtosis for the three groups is -0.823, 0.083, and 7.74; therefore  the violation is considered minor except for one case and the data are assumed to follow a normal distribution. Levene's test showed that all eight data sets did not violate the homogeneity of variances assumption.

A significant main effect for adaptation level was obtained $F(1,25) = 2402.87, p<0.001, \eta_p^2 =0.990$ with confidence levels the success rate increased from $0.68 \pm 0.020$ for slow-adapters to $0.80 \pm 0.023$ for fast-adapters. A significant main effect for sessions was obtained, $F(1,25) = 14.93, p <0.001, \eta_p^2 =0.374$ with confidence levels, the success rate increased from $0.69 \pm 0.025$ in Session-1 to $0.79 \pm 0.014$ in Session-2. A significant main effect for task constraints was obtained, $F(1,25) = 48.36, p<0.001, \eta_p^2 =0.659$ with confidence levels, the success rate decreased from $0.86 \pm 0.02$ in low-constraint tasks to $0.62 \pm 0.026$ in high-constraint tasks. The interaction between adaptation level and sessions did not show a significant effect, $F(1,25) = 3.15, p = 0.088, \eta_p^2 =0.112$. However, examination of the marginal means indicated that slow-adapters have an improvement in their success rate from $0.61 \pm 0.033$ in Session-1 to $0.76 \pm 0.019$ in Session-2, while fast-adapters have a slight improvement from $0.78 \pm 0.037$ to $0.83 \pm 0.021$. 

\begin{figure}[t]

\centering
\includegraphics[width=0.5\textwidth]{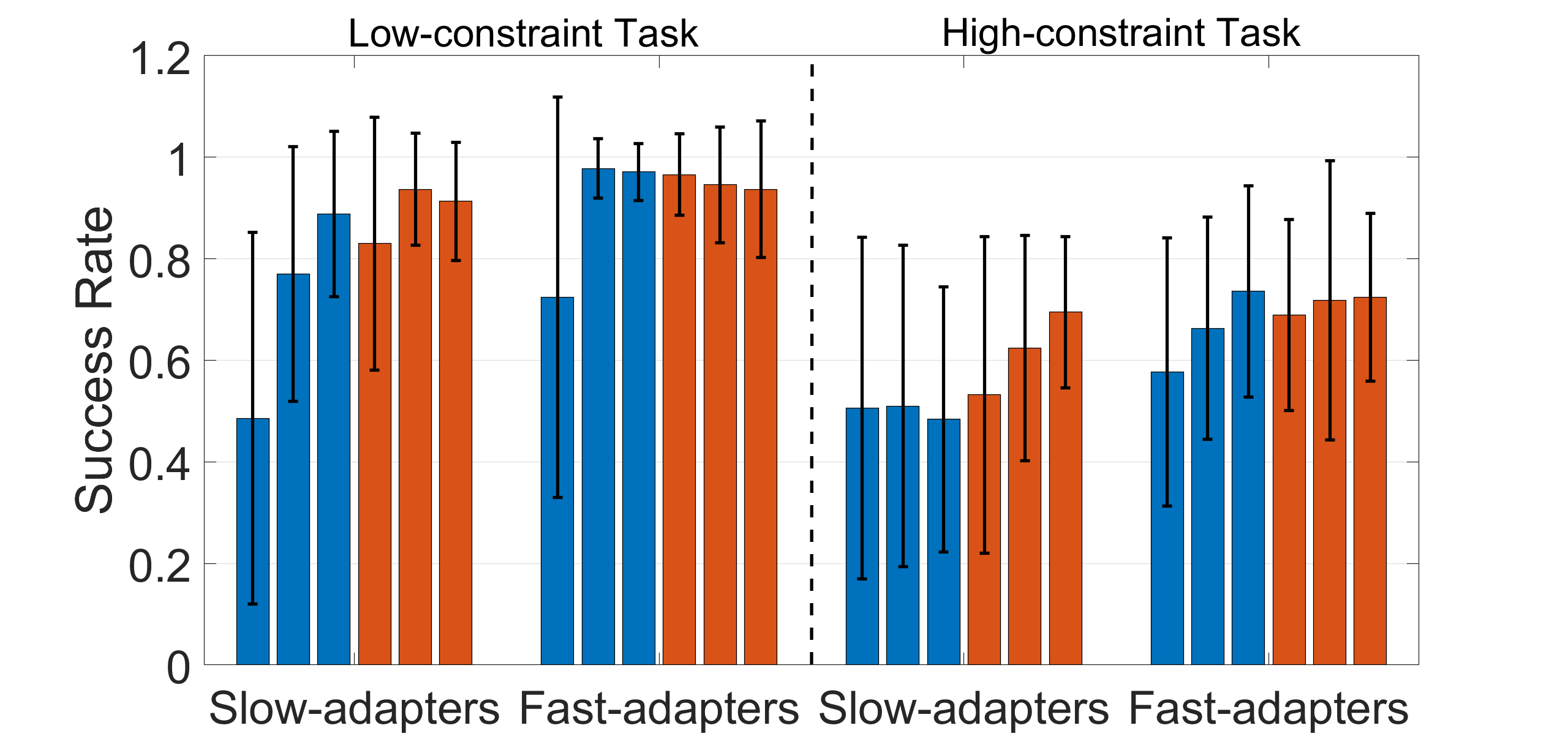}

\caption{Generalization success rate for both slow- and fast-adapters  in all trials. Blue bars represent Session-1 trials and red bars represent Session-2 trials.}
\vspace*{-3mm}
\label{fig:extrapolation_trials}
\end{figure}

Examining the individual trials of both sessions, we found a similar trend as with learned task performance, with slow-adapters taking longer to reach high performance than the fast-adapters as shown in Figure~\ref{fig:extrapolation_trials}. Particularly in the high-constraint tasks, slow-adapters have a steady low performance in the first session's trials with $0.51 \pm 0.34$ and then gradually improve in the second session until they reach $0.70 \pm 0.15$. Fast-adapters demonstrate gradual improvement in Session-1 trials and almost converge to high performance at the end of their first session with $0.74 \pm 0.21$. For the low-constraint tasks, slow-adapters converge to their highest performance in the second trial of Session-2 with $0.94 \pm 0.11$ while fast-adapters converge to a higher performance of $0.98 \pm 0.06$ than slow-adapters in the second trial of Session-1.

\section{DISCUSSION}
        \label{sec:discussion}
        The results show a strong correlation between task performance and generalization. In addition,a clear difference between fast- and slow-adapters in both task and generalization performance. In addition, practice makes a significant difference in slow-adapters' performance, approaching fast-adapters' performance. This finding agrees with our previous work demonstrating the need for training for slow-adapters to provide high-quality demonstrations for robots~\cite{sakr2020training}. 

The categorization of users into fast-adapters and slow-adapters showed that the stated robotics expertise may not be reflected in the task performance and generalization. The adaptation level to the task is greatly affected by the previous expertise of each individual~\cite{kobak2012adaptation}. This suggests that using only the questionnaire may not be sufficient as skills from other domains could be transferable to the robotics domain and contribute to a participant's performance and adaptation~\cite{everson1997metacognitive}.

Results show support for \textbf{H1} with a significant correlation between task performance and generalization. Task learning results show support for \textbf{H2} with a significant main effect of the adaptation level on the success rate. Similarly, support for \textbf{H3} was found as there is a significant difference between both sessions for slow-adapters with 43\% improvement between sessions, while fast-adapters have only 11\% improvement. The importance of practice for slow-adapters was shown with the significant difference between fast- and slow-adapters in Session-1 with an estimated mean difference of $0.21 \pm 0.055$, while no significant difference between them was observed in Session-2, the mean difference was $0.04 \pm 0.035$. 

Another interesting finding is that the difference between the two sessions in the high-constraint task is double the difference in the low-constraint task. The reason is the greater room for improvement in the high-constraint task while the low-constraint task is straightforward and does not require the same amount of practice or expertise. This suggests taking task complexity into consideration when providing training for slow-adapters before teaching robots by demonstration.  

Task generalization results show a similar pattern as the task learning results, providing support for \textbf{H4}, since fast- and slow-adapters are significantly different. Partial support for \textbf{H5} was found. Although a non-significant interaction between sessions and adaptation was detected, the estimated marginal means increased by 25\% from Session-1 to Session-2. Furthermore, the difference between fast- and slow-adapters in the first session was $0.19 \pm 0.049$, while in Session-2 this difference dropped to $0.08 \pm 0.028$. Unlike task learning results, the improvement between sessions in the low-constraint task was $0.12 \pm 0.032$, while in the high-constraint task it was $0.09 \pm 0.047$. This is because extrapolation in the constrained space is challenging, especially for the farthest points on the box's face. 


One way to explain these conclusions is through the motor skill learning literature. In \cite{fitts1967human}, Fitts and Posner describe a model for skill acquisition. Their model proposes three stages for acquiring a motor skill: the cognitive stage, associative stage and autonomous stage. In the first stage, a trainee processes the received information about the task and tries out several strategies to perform it. This results in a high cognitive load as well as highly variable performance. After finding a good strategy, the trainee moves to the associative stage at which he/she tries to refine the strategy to improve the task performance. In the last stage, the autonomous stage, the trainee can perform the task more accurately with less cognitive load. 

Applying this model to the studies in this paper, it is clearly shown in Figure~\ref{fig:interpolation_trials} and Figure~\ref{fig:extrapolation_trials} that both fast- and slow-adapters achieve low performance in the very first trial (cognitive stage) with high variance. This is because of the high cognitive load for users to define the strategy for demonstrating the task to the robot. This suggests that the first demonstrations for a task should not be considered for robot learning regardless of the user's stated expertise level. This reasoning aligns with~\cite{cakmak2014eliciting}, who find that  users are not spontaneously optimal in providing demonstrations. Furthermore, fast-adapters progress faster to associative and autonomous stages while slow-adapters take a longer time. This was shown by the improvement in Session-2 compared to Session-1 for slow-adapters, while fast-adapters show a relative consistency in their success rate right after the very first trial.

\section{CONCLUSION}
        \label{sec:conclusion}
        In this paper, we proposed a framework for defining the quality of the provided demonstrations using task learning and generalization performance. This proposed framework was validated in a generic learning task with two levels of constraints. The demonstrations data were collected in two sessions on two different days to determine the improvement pattern in robot teaching skills for both slow- and fast-adapters. The results show a significant correlation between task performance and generalization performance. In addition, a significant difference between slow- and fast-adapters was shown in both task constraints. Furthermore, slow-adapters show a significant improvement from the first session to the second one and even get close to the performance of the fast-adapters in the second session. Fast-adapters show consistent performance between the two sessions as they defined the strategy for approaching the task faster than slow-adapters. We showed how these results are consistent with the motor skills learning literature. 



\bibliographystyle{IEEEtran}
\bibliography{bibliography}

\end{document}